\documentclass[runningheads]{llncs}
\usepackage[T1]{fontenc}
\usepackage[caption=false]{subfig}
\usepackage{graphicx}
\usepackage{amsmath}
\usepackage{amssymb}
\usepackage{booktabs}
\usepackage{adjustbox}
\usepackage{hyperref}
\begin{document}
\title{Not with my name! Inferring artists' names of input strings employed by Diffusion Models}

\author{Roberto Leotta\inst{1}\orcidID{0000-0003-4441-8313} \and
Oliver Giudice\inst{2}\orcidID{0000-0002-8343-2049}\and \\
Luca Guarnera\inst{3}\orcidID{0000-0001-8315-351X} \and \\
Sebastiano Battiato\inst{1,3}\orcidID{0000-0001-6127-2470}}

%
%\authorrunning{R. Leotta et al.}
\institute{iCTLab Spinoff of University of Catania, Italy\and
Applied Research Team, IT dept., Banca d'Italia, Italy \and %95125 Catania\\
Department of Mathematics and Computer Science, University of Catania, Italy \\
\email{roberto.leotta@ictlab.srl},  \email{oliver.giudice@bancaditalia.it}, \\
\email{\{luca.guarnera, sebastiano.battiato\}@unict.it}}

\maketitle              % typeset the header of the contribution
\begin{abstract}
Diffusion Models (DM) are highly effective at generating realistic, high-quality images. However, these models lack creativity and merely compose outputs based on their training data, guided by a textual input provided at creation time. Is it acceptable to generate images reminiscent of an artist, employing his name as input? This imply that if the DM is able to replicate an artist's work then it was trained on some or all of his artworks thus  violating copyright. In this paper, a preliminary study to infer the probability of use of an artist's name in the input string of a generated image is presented. To this aim we focused only on images generated by the famous DALL-E 2 and collected images (both original and generated) of five renowned artists. Finally, a dedicated Siamese Neural Network was employed to have a first kind of probability. Experimental results demonstrate that our approach is an optimal starting point and can be employed as a prior for predicting a complete input string of an investigated image. Dataset and code are available at: \href{https://github.com/ictlab-unict/not-with-my-name}{https://github.com/ictlab-unict/not-with-my-name}.

\end{abstract}

\section{Introduction}
\label{sec:intro}

The rapid advancement of generative models, particularly Diffusion Models~\cite{dhariwal2021diffusion}, has led to a surge in high-quality, realistic image generation. These models have demonstrated immense potential for creative applications across various domains, including art, design, and advertising. However, their ability to replicate the styles of specific artists raises concerns about Intellectual Property (IP) rights and potential copyright infringements~\footnote{\href{https://www.theverge.com/2023/1/16/23557098/generative-ai-art-copyright-legal-lawsuit-stable-diffusion-midjourney-deviantart}{https://www.theverge.com/2023/1/16/23557098/generative-ai-art-copyright-legal-lawsuit-stable-diffusion-midjourney-deviantart}} \cite{vyas2023provable,zirpoli2023generative,shan2023glaze}. 

As the boundary between human creativity and machine-generated content becomes increasingly blurred, it is crucial to address the legal and ethical implications of using generative models to produce art, as well as to develop methods that evaluate the extent to which generated images are influenced by the works of real artists and ensure the protection of their intellectual property.

Several studies and articles have explored the legal ramifications of generative models and their potential to infringe on copyrights~\footnote{\href{https://www.theverge.com/23444685/generative-ai-copyright-infringement-legal-fair-use-training-data}{https://www.theverge.com/23444685/generative-ai-copyright-infringement-legal-fair-use-training-data}}. % already highlighted instances where generative models have been used to create content that closely resembles copyrighted material, thereby raising questions about the fair use of training data and the ownership of generated content[3]. 
In an article dated 2021 the challenges in determining copyright ownership for AI-generated works were discussed and authors emphasized the need for legal frameworks that could adequately address the unique nature of generative models~\footnote{\href{https://www.oreilly.com/radar/what-does-copyright-say-about-generative-models/}{https://www.oreilly.com/radar/what-does-copyright-say-about-generative-models/}}. MLQ.AI also reported on a copyright infringement case involving generative AI~\footnote{\href{https://www.mlq.ai/copyright-infringement-generative-ai-this-week-in-ai/}{https://www.mlq.ai/copyright-infringement-generative-ai-this-week-in-ai/}}, which sparked debates on the responsibilities of AI developers and users in protecting original creators' rights. %Data Science Central (2021) provided a comprehensive analysis of copyright protection and generative models, discussing the implications of AI-generated content on current legal frameworks[6].
In response to these concerns, legal scholars have delved into the complexities of copyright law as it pertains to AI-generated artworks. Gillotte~\cite{gillotte2019copyright} examined the challenges in assigning liability and protecting IP rights, arguing that existing copyright laws may not be sufficient to address the unique characteristics of AI-generated contents. Indeed, it is became a copyright dilemma and the need for a balance between innovation and IP protection is arising to ensure that creative works are safeguarded without stifling technological advancements~\cite{hristov2016artificial}.

It is not easy to determine how images generated by tools like DALL-E 2 or Midjourney are created by combination of images employed at training time. But, it could useful to develop tools able to deduce the textual prompts that generated an investigated image. Recently, a Kaggle competition was launched on the task~\footnote{\href{https://www.kaggle.com/competitions/stable-diffusion-image-to-prompts/data}{https://www.kaggle.com/competitions/stable-diffusion-image-to-prompts/data}} but it still lacks of effective methods. However, to make images that appear with an artist's style the prompt of the generating tools should necessary contain the artist's name. Indeed, the generating tool was trained with original artworks belonging to that artist, coupled (as labels) with sentences containing the artist's name. This should demonstrate at a certain level the use of artist's artworks at training time thus violating copyright. Moreover, also the use of the name of a person without his consent should arise other issues.

In order make a starting point in the state of the art for techniques able to protect artists, in this study, an introductory empirical analysis method is presented to infer if an artist's name was employed to generate an investigated image. To simplify the problem, an extremely constrained scenario was built by collecting a dataset composed of original artworks from five renowned artists and also images generated employing the OpenAI's DALL-E 2~\cite{ramesh2022hierarchical} with artists' names employed as prompted textual strings. Starting from these data a Siamese Neural Network was exploited to learn a dedicated metric for the purpose.
Through a series of experiments, the effectiveness of the proposed approach was demonstrated in identifying a sort of probability of the usage of an artist's name in the generation process of an image. This study can be extremely important in forensics in order to reconstruct and analyze the history of multimedia content~\cite{battiato2016multimedia}. %Results, code and datasets will be publicly available on a dedicated website and on Github at camera-ready.%the extent to which generated images are influenced by the reference dataset. We also explore the potential of our Siamese Network-based metric to serve as an alert system for potential IP infringements, prompting further analysis and comparison by experts. By providing insights into the relationship between generative models and copyrighted material, our research seeks to inform the development of legal frameworks and best practices that strike a balance between innovation, artistic expression, and the protection of intellectual property rights.

The remainder of this paper is organized as follows: Section~\ref{ref:sota} lists research papers on the topic introducing the reader to the ethical and IP problems; Section~\ref{ref:dataset} presents the collected dataset with details on composition and how it was built; Section~\ref{ref:approach} describes the proposed approach starting from a discriminative solution (Section~\ref{ref:resnet-discriminating}) to the final metric objective of the study (Section~\ref{ref:siamese-metric}). Experimental results are presented and discussed in Section~\ref{ref:results} and Section~\ref{ref:conclusion} concludes the paper with some hints for future development.

\section{Related Works}
\label{ref:sota}

Generative models are a type of machine learning model that use data generation techniques to create new data instances from an existing dataset. These models can be used to generate images, text, sound and other types of data, and have multiple applications in fields such as content creation, simulation and new product creation. Mainly there are two large families of modern generative techniques: the Generative Adversarial Networks (GAN)~\cite{goodfellow2014generative} and the Diffusion Models (DM)~\cite{dhariwal2021diffusion}. The latter are ultimately being employed by successful applications given their simplicity with which they allow the user to control how the multimedia content has to be generated.

Regarding the intellectual property of the data on which generative models have been trained, it can be an important issue as these models can be trained on large amounts of data that belong to other owners, such as images or texts on the internet. If so, the use of this data by generative models could be considered an infringement of copyright or other forms of intellectual property~\cite{abbott2022intellectual}.

Furthermore, since generative models use training data to learn and create new data, they can also create potential cybersecurity and privacy issues. For example, if a generative model was trained on individuals' personal data, it could generate new personal information that could be used for improper purposes.

Given the potential risks of generated multimedia contents, several state of the art works already addressed the issue by trying to detect if a query sample is real or fake~\cite{zhang2019detecting,wang2019fakespotter}) or even recognize the specific GAN architecture that generated it~\cite{Giudice_JI_2021,wang2020cnn,guarnera2020fighting,guarnera2022face}. Contents generated by DMs are already being investigated and there are some works that try to expose them~\cite{sha2022fake,corvi2022detection}. 

However, detecting a synthetic content is just the first step. When it comes to evaluate if a certain image infringes IP property different approaches have to be developed. Yet while the problem is only discussed in newspapers first research papers are being published.

The most interesting one is the work of Carlini et al~\cite{carlini2023extracting} which analyses different DMs in order to infer if a certain image was employed in the training set given a set of constraining hypothesis. This is a great starting point giving hints on possible discriminating features, however it only addresses privacy issues. 

A method to effectively protect the IP is yet to be proposed. This work is a first attempt at predicting if a specific text (i.e. artists's names) was employed as input prompt at generation time.

\section{A dataset of original and synthetic artworks}
\label{ref:dataset}

%\color{blue}
Three different dedicated datasets were collected to evaluate the effectiveness of modern Diffusion Models in generating images of specific artists and to have a basis to train the solutions proposed in this paper. %Below is a detailed description of each dataset.

%In this study, we used three different datasets to evaluate the effectiveness of modern diffusion models in generating images of specific artists and to analyze the implications for intellectual property rights. Below is a detailed description of each dataset.

%\subsection{Synthetic Image Dataset}
%\label{ref:dataset-synthetic}
%To generate the synthetic images, we selected five prominent artists: 

The first two datasets, containing synthetic and original data respectively, were created as described below.

Five prominent artists were selected to generate the synthetic images: Alfred Sisley, Claude Monet, Pablo Picasso, Paul Cezanne and Pierre-Auguste Renoir. 
For each artist, $2,000$ synthetic images with a size of $512 \times 512$ pixels, were generated using DALL-E 2 and employing $1,000$ different descriptive text lines similar to the following:
%For each artist, we created $2,000$ synthetic images with a size of $512 \times 512$ using DALL-E 2. We generated these images based on $1,000$ descriptive text lines as:
%. Some examples of the text lines before adding the artist's name are:
%\begin{itemize}
%\item 
\textit{A field of sunflowers with a blue sky background};
%\item 
\textit{A group of children playing at a playground in a city park};
%\item 
\textit{A group of elephants drinking at a watering hole in the savannah}.

%\end{itemize}

To ensure that the generated images had the same style as the selected artist, the text "\textit{by Artist's name}" was added to each line of input prompting text, such as follows:
%To ensure the selected artist's style was maintained, we added the prefix "by Artist's name" to each text line: %. For example:
%\begin{itemize}
%\item 
\textit{A field of sunflowers with a blue sky background, \textbf{by Pablo Picasso}};
%\item 
\textit{A group of children playing at a playground in a city park, \textbf{by Alfred Sisley}};
%\item 
\textit{A group of elephants drinking at a watering hole in the savannah, by \textbf{Claude Monet}}.
%\end{itemize}

In this way, two images per artist were generated for each text line. In addition, different context sentences were used for each artist, thus ensuring a diverse representation of the artist's style across subjects and maximizing variability.
%For each text line, we generated two images per artist, resulting in a total of $10$ synthetic images (two images per artist, for five artists). Therefore, each text line was used for each artist, ensuring a diverse representation of the artist's style across different subjects.

\begin{table*}[t!]
\centering
\caption{Number of images employed. From the second to the last column the artists' names are given. The last two rows describe the total number of synthetic and real original images collected for each artist.}
\begin{adjustbox}{max width=\textwidth}
\begin{tabular}{cccccc}
\cline{2-6}
                    & \textbf{Alfred Sisley} & \textbf{Claude Monet} & \textbf{Pablo Picasso} & \textbf{Paul Cezanne} & \textbf{Pierre-Auguste Renoir} \\ \hline
\textbf{\#Synthetic} & 1,722                  & 1,702                 & 1,515                  & 1,734                 & 1,846                          \\ \hline
\textbf{\#Original}      & 470                    & 1,044                 & 1,019                  & 588                   & 1,008                          \\ \hline
\end{tabular}
\end{adjustbox}
\label{tab:dbNumbers}
\end{table*}

In order to work with synthetic data representing the style of the involved artists, validation and cleaning operations of the generated dataset were necessary because the DALL-E 2 algorithm also produces images that did not match the intended artistic style or content (e.g., photo-realistic images rather than paintings).

In addition, original paintings of the same five artists were also downloaded from WikiArt~\footnote{\url{www.wikiart.org}}. 
Table~\ref{tab:dbNumbers} shows the total number of original and synthetic images used in the experimental phase. The original images have different resolutions and, as shown in the Table~\ref{tab:dbNumbers}, some artists have a limited number of works.

%Validation and cleaning of the synthetic dataset were necessary because the DALL-E 2 generation algorithm occasionally produced images that did not match the expected artistic style or content. For instance, some generated images were photorealistic rather than paintings, which did not align with our goal of obtaining artist-specific paintings. By manually reviewing the generated images, we ensured that the dataset only contained images that accurately represented the style and content of the selected artists.
%After the validation and cleaning process, we obtained the following number of synthetic images for each artist:
%\begin{itemize}
%\item Alfred Sisley: $1722$ synthetic images;
%\item Claude Monet: $1702$ synthetic images;
%\item Pablo Picasso: $1515$ synthetic images;
%\item Paul Cezanne: $1734$ synthetic images;
%\item Pierre-Auguste Renoir: $1846$ synthetic images.
%\end{itemize}

%\subsection{Real Image Dataset}
%\label{ref:dataset-real}

%On the other hand, we created a real image dataset by downloading the artworks of the same five artists from WikiArt \cite{www.wikiart.org}. The number of real images for each artist is as follows:
%\begin{itemize}
%\item Alfred Sisley: $470$ real images;
%\item Claude Monet: $1044$ real images;
%\item Pablo Picasso: $1019$ real images;
%\item Paul Cezanne: $588$ real images;
%\item Pierre-Auguste Renoir: $1008$ real images.
%\end{itemize}
%The real images have varying resolutions, and as can be seen, some artists have a limited number of completed works available.

%\subsection{Mixed Dataset}
%\label{ref:dataset-mix}
 
Finally, a third mixed dataset was created, consisting of $50\%$ original images and $50\%$ synthetic images. To obtain a balanced dataset, $470$ images from the synthetic image dataset and $470$ images from the real image dataset for each artist were randomly selected. This choice was made as Alfred Sisley has $470$ images in the real image dataset.

The three datasets were split into two parts: $80\%$ as training set and the remaining $20\%$ of samples as validation set. %All the described datasets, along with the text lines used to generate the synthetic image dataset, will be available for download on the project's GitHub page at acceptance time.

%\color{black}

\section{The proposed approach}
\label{ref:approach}

\begin{figure*}[t!]
    \centering 
    \includegraphics[width=\linewidth]{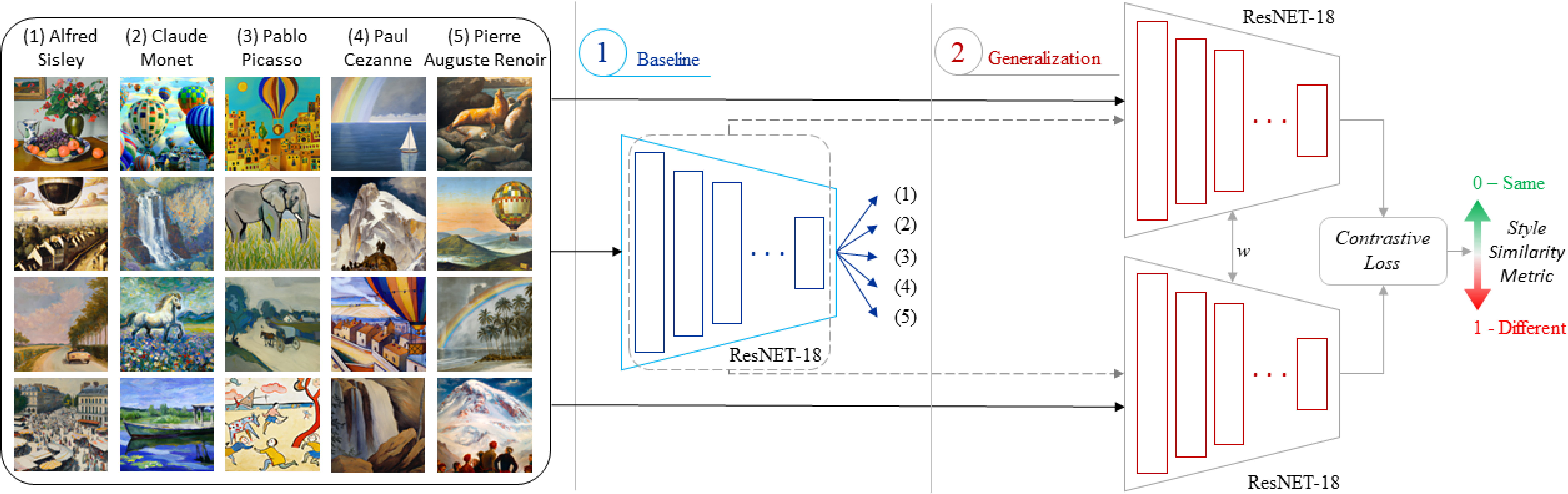}
    %\vspace{-0.5cm}
    \caption{Proposed approach. The dataset of artists (real and synthetic) was used in the Baseline approach (1) to discriminate against artist. To achieve more generalization, explainability and best define the similarity between different artist styles (2), a Siamese engine was trained from the pretrained model of (1). The ResNET-18 architecture was used in all experiments.
    }
    \label{fig:proposedpipeline}
\end{figure*}
The ultimate goal of this study was to develop a metric, a sort of probability that a content generated by OpenAI's DALL-E 2 was obtained prompting the tool with a sentence containing the name of one of the artists to be protected.

As a starting point, let $D$ be the \textit{reference dataset} to be protected containing a finite number of authors and their artworks. The presented metric will evaluate the query image by means of comparison with all the elements in $D$. Specifically, the objective is to estimate a set of distances $L = \widehat{d}(I,s)$ where $I$ is the image to be investigated and $s \in D$ is an original image of the artists taken into account. The \textit{$\widehat{d}$} function is not known and has to be modelled in such a way that \textit{L} has to be close to 0 if the two considered samples belong to the same author (both being original or synthetic and generated with his name). On the contrary, values of $L$ should be 1 or bigger if the two considered samples are unluckily related. 

%\textbf{qui bisogna definire il problema}
%\textbf{da ICIP}
%\color{green}
%The proposed Forensics Ballistics Analysis is carried out by means of comparison: a selected cartridge is compared to another cartridge found on a crime scene. The goal is to estimate the value $L = \widehat{d}(s_1,s_2)$ where $s_1$ and $s_2$ are two cartridge samples with their own impressions left by an unknown firing pin of a gun and \textit{$\widehat{d}$} is a function to be modelled for the correct estimation of the value \textit{L}. The function \textit{$\widehat{d}$} is usually modelled to range in $[0,1]$. Specifically, \textit{L} has to be close to 0 if the two considered samples were fired by the same gun (i.e. the cartridge can be considered a forensics evidence). On the contrary, value of $L$ should be close to 1 if the two considered samples were fired by different guns. Also, $L$ should be small in case two samples were fired by guns of the same model. 
%\color{red}

At first, an evaluation study of a method to classify images between authors with the aim to learn the discriminative features was carried out. Further analysis was then carried out in order to model the final metric by means of a Siamese Neural Network (taking inspiration from \cite{giudice2019siamese,guarnera2019new,battiato2022cnn} with the objective of learning a distance \textit{$\widehat{d}$}. The overall study could be schematized as described in Figure~\ref{fig:proposedpipeline}.

\subsection{Learning to discriminate}
\label{ref:resnet-discriminating}

%\begin{figure}[t!]
%    \centering 
%    \includegraphics[width=1.1\linewidth]{images/2_train_val_acc.png}
%    %\vspace{-0.5cm}
%    \caption{Training and validation accuracy of the Resnet-18 for authors discrimination trained on %the original images. Orange shows the trend on the training set while blue is the validation one.}
%    \label{fig:transfer_train}
%\end{figure}

\begin{figure}
\centering
     \subfloat[\label{fig:conf_transfer}]{
     \centering
       \includegraphics[scale=0.3]{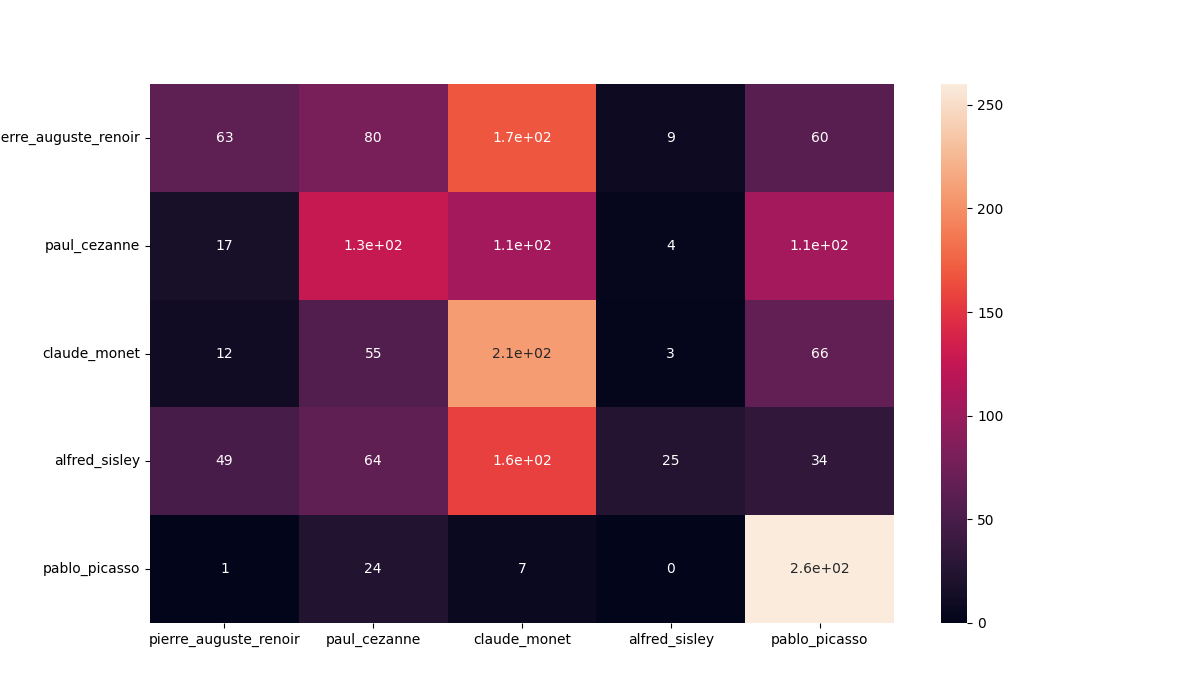}
     }
     \hfill
     \subfloat[\label{fig:conf_all}]{%
     \centering
       \includegraphics[scale=0.3]{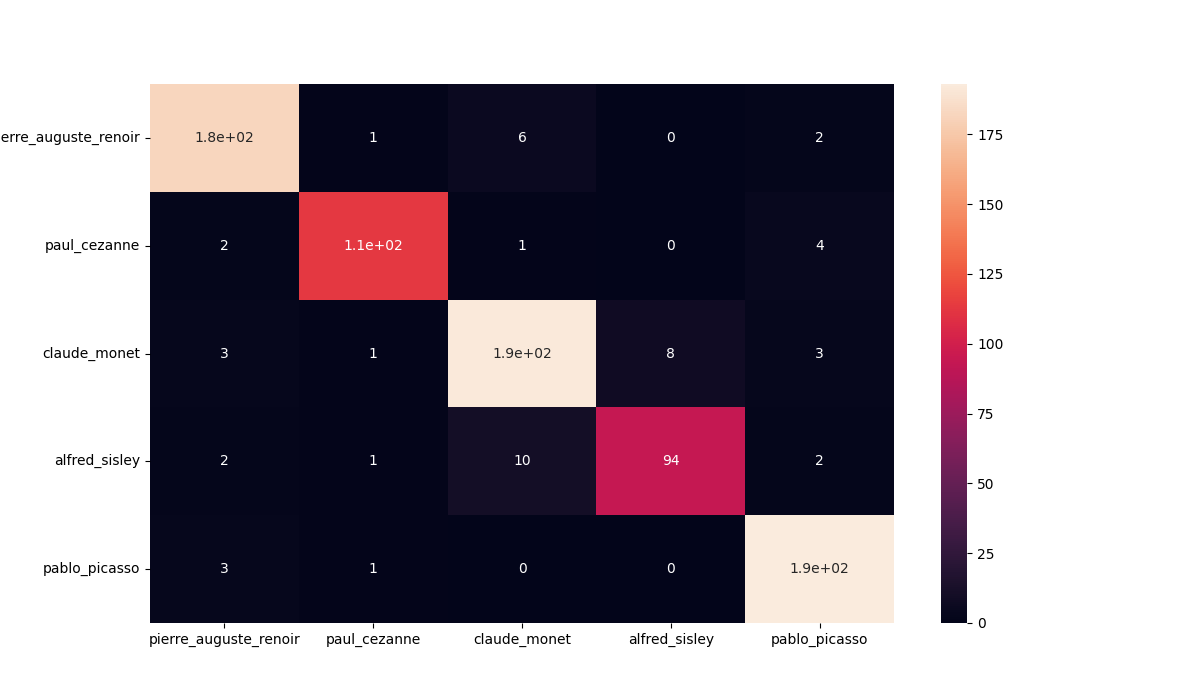}
     }
    \hfill
    %\vspace{-0.3cm}
     \caption{(a) Transfer learning attempt. Results are shown as a confusion matrix computed on synthetic images with the Resnet-18 model trained only on original artworks. (b) Confusion matrix obtained on test set with the Resnet-18 model trained only on mixed dataset containing both original and synthetic images.}
     \label{fig:conf}
\end{figure}

%\begin{figure}[t!]
%    \centering 
%    \includegraphics[width=\linewidth]{images/1_model_validation_su_sintetiche_confusion_matrix.png}
    %\vspace{-0.5cm}
 %   \caption{Transfer learning attempt. Results are shown as a confusion matrix computed on synthetic images with the Resnet-18 model trained only on original artworks.}
  %  \label{fig:conf_transfer}
%\end{figure}

%\begin{figure}[t!]
 %   \centering 
  %  \includegraphics[width=\linewidth]{images/2_test_confusion_matrix.png}
    %\vspace{-0.5cm}
   % \caption{Confusion matrix obtained on test set with the Resnet-18 model trained only on mixed dataset containing both original and synthetic images.}
   % \label{fig:conf_all}
%\end{figure}

In order to demonstrate the possibility of discriminating the authors considered in our dataset $D$. Taking inspiration from~\cite{sha2022fake,corvi2022detection}, a Resnet-18~\cite{he2016deep} was employed.% for this task.

The first training experiment was to assess the Resnet-18 ability to discriminate between the different artists employing the original artworks only as training data. This set of images was split into 80\% for training and 20\% for testing purposes. After training the Resnet-18 model, excellent results in discriminating the artists were obtained on the original images meaning that the model successfully captured the distinctive characteristics of each artist's style, which was reflected in the high accuracy achieved during testing.% as shown in Figure~\ref{fig:transfer_train}. 
However, the trained Resnet-18 model applied on the synthetic images, was not able to obtain similar good results as shown in Figure~\ref{fig:conf}(a). The model performances significantly deteriorated, with only a few artists being discriminated with reasonable accuracy. This indicated that the model struggled to generalize its understanding of the artists' styles for the synthetic images. It is clear that this \textit{transfer learning} approach was not able to take into account features extractable from synthetic images, probably because of the generative process itself \cite{guarnera2020fighting}. %Moreover, results somewhat reflect the number of actual artworks available for each author. Alfred Sisley, which has fewer known and publicly available artworks than the others, shows fewer correct predictions. This is an interesting insight: DALL-E 2 was probably trained on a set of original artworks with the same distribution and cardinality as in the employed dataset $D$.

Given that the before-mentioned transfer learning attempt was not able to correctly discriminate synthetic images, a new Resnet-18 model was trained with the mixed dataset described in Section~\ref{ref:dataset} (with the common train-test split 80\%-20\% as well). This time, the results obtained were way better then previous one so it can be said that this last Resnet-18 model learned not only features about authors' styles but also how DALL-E 2 alters images invisibly (i.e. by producing different distributions of frequencies of the generated images with respect to original ones). Upon analyzing the confusion matrix (Figure~\ref{fig:conf}(b)), it was possible to find an higher entropy for those artists with fewer publicly available works in the dataset. This could give an hint about DALLE-E training phase: it might have been trained on a dataset with a similar distribution and cardinality as the employed dataset $D$, which led to the generative model's inability to specialize in the artistic styles of specific artists with limited available samples. Obviously this is a strong assumption given that it is a just an introductory, trivial and preliminary insight, which could lead to further investigation on the composition of the dataset employed by the DM.

Given the obtained results, for the final objective of this paper this last model will be employed for further investigations described in the next Section.

%This initial experiment served as a baseline to understand the model's performance when trained only on real images and tested on synthetic images. The subsequent experiment, where the model was trained on the mixed dataset, was designed to address these limitations and improve the model's performance on both real and synthetic images.

%At first only the original artworks were considered. A common train-test split (80\%-20\%) was applied to this set of data. The trained network performed well with good level of accuracy in discriminating authors. A further test was carried out with this trained model to discriminate the set of images generated with DALL-E 2, obtaining results shown in Figure \ref{fig:conf_transfer}. It is clear that this transfer learning approach is not taking into account features that could be detectable on generated images, probably because all of those alterations introduced by the generative process \cite{guarnera2020deepfake}. Moreover, results somewhat reflect the number of actual artworks available for each author. Alfred Sisley, which has fewer known and publicly available artworks than the others, shows fewer correct predictions. This is an interesting insight: DALL-E 2 was probably trained on a set of original artworks with the same distribution and cardinality as in the employed dataset $D$.

\subsection{Learning a metric for mixed artworks}
\label{ref:siamese-metric}

While the goal of a single Resnet-18 was to learn a hierarchy of feature representations to solve the discriminative tasks between 5 classes (the five considered artists), a Siamese  Neural Network can be exploited for weakly supervised metric learning tasks. Instead of taking single sample as input, the network takes a pair of samples, and the loss functions are usually defined over pairs. In this case the loss function of a pair has the following form:
%\vspace{-0.2cm}
\begin{equation}
\label{eq:loss}
	L(s_1, s_2, y) = (1 - y) \alpha D^2_w + y \beta e^{\gamma D_w}
\end{equation}
where $s_1$ and $s_2$ are two pair of samples, $y \in {0,1}$ is the similarity label, and $D_w$ is a distance function defined as in \cite{wang2015sketch}. Parameters were set as follows $\alpha = 1 / C_p$, $\beta = C_n$ and $\gamma = -2.77/C_n$ where $C_p = 0.2$ and $C_n = 10$.

Unlike methods that assign binary similarity labels to pairs, the network aims at bringing the output feature vectors closer for input pairs of the training labeled as similar, or push the feature vectors away if the input pairs are labeled as dissimilar. 

\begin{figure}[t!]
    \centering 
    \includegraphics[width=\linewidth]{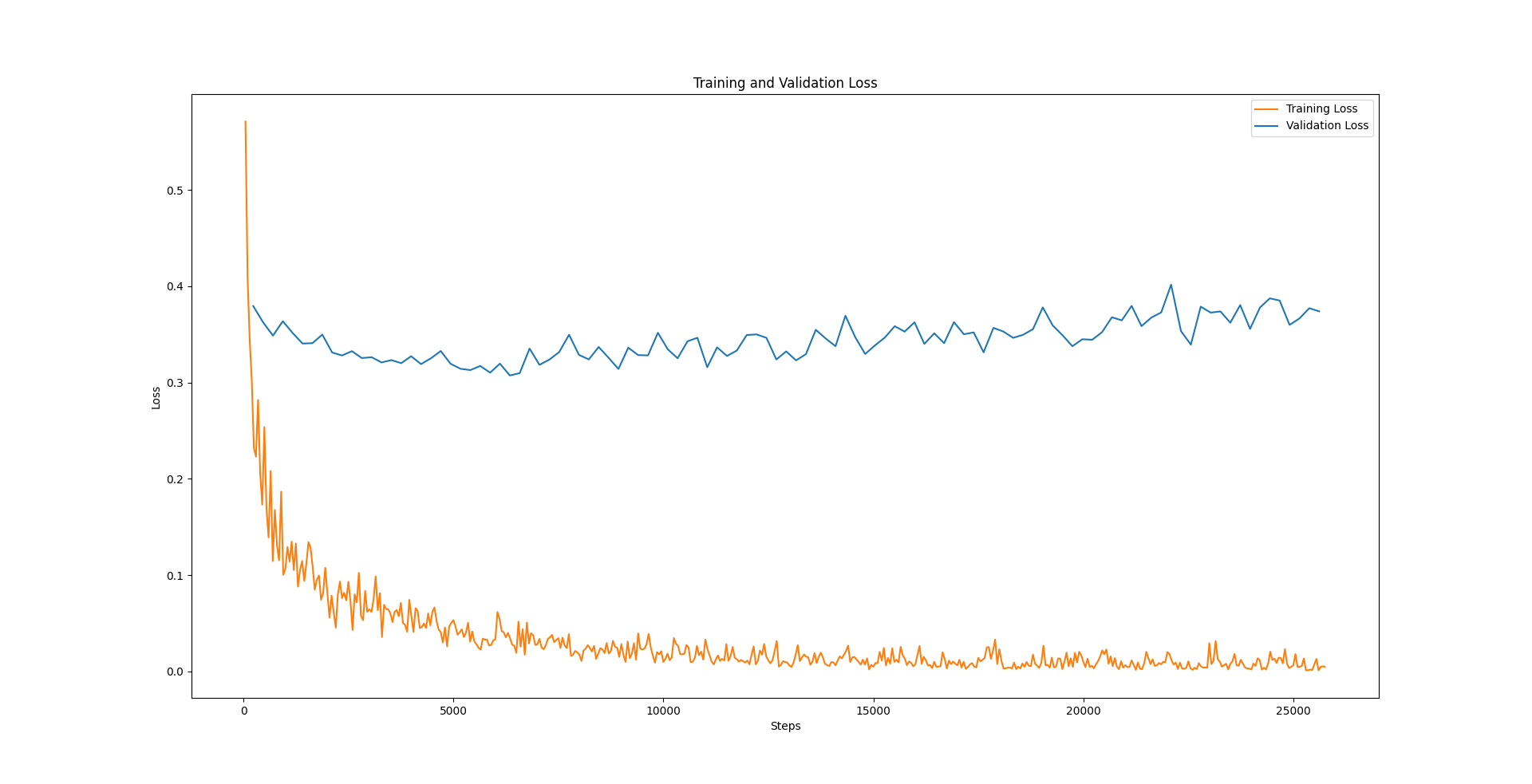}
    %\vspace{-0.5cm}
    \caption{Training losses of the siamese architecture. Orange shows the trend on the training set while blue is the validation one.}
    \label{fig:siamese_train_loss}
\end{figure}

Figure \ref{fig:proposedpipeline} shows the overall architecture showing as baselines the Resnet-18 models as trained and described in previous Section. The siamese network was trained employing the same training set used to previously train the single Resnet-18 models from the mixed dataset. Training and validation loss are shown in Figure \ref{fig:siamese_train_loss}. Best model was considered for experiments with respect to the elbow of validation loss as shown in Figure \ref{fig:siamese_train_loss}.

\begin{figure*}[t!]
    \centering 
    \includegraphics[width=\linewidth]{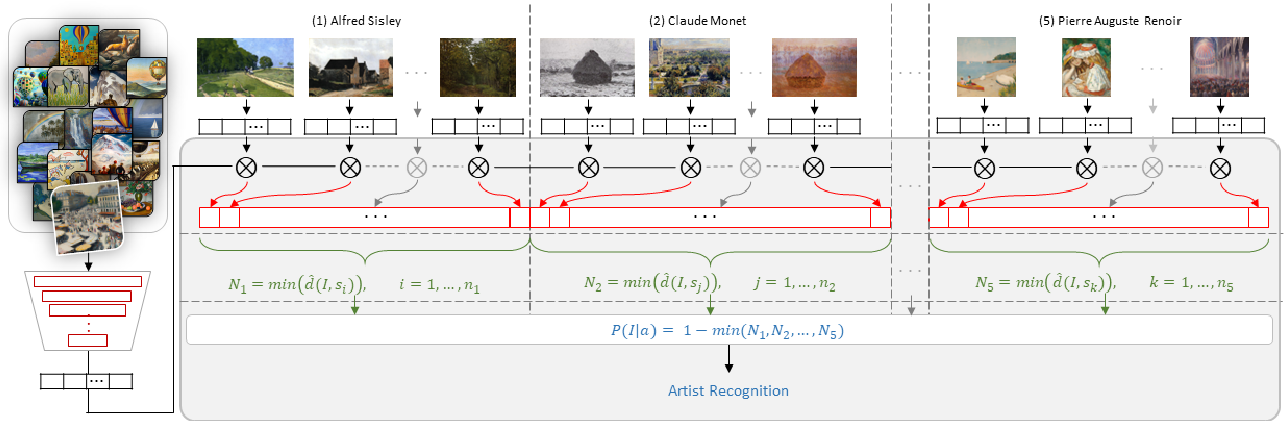}
    %\vspace{-0.5cm}
    \caption{Final approach. Given an input image to ResNET-18, the similarity score (defined by the Siamese approach) with respect to each real image of the involved artists is calculated. A voting process will define the number of images for each artist whose similarity score S exceeds a set threshold value T. Then, the test image will be assigned to a specific artist with respect to the highest score obtained in the previous step.
    }
    \label{fig:siamese_final}
\end{figure*}

Given the properties of the loss $L$ learned by the siamese network in a context of balances classes, the trained distance $\widehat{d}$ is a likelihood function describing the possibility to have image $I$ being generated by one of the five authors' names took into account\cite{berlemont2018class}. Thus, it is possible to define the probability that image $I$ was generated by the tool DALL-E 2, by prompting a string containing an artist's name $a$ between the five considered:

\begin{equation}
\label{eq:prob}
    P (I|a) = 1 - min(\widehat{d}(I,s_a)) 
\end{equation}
where $\widehat{d}$ is the distance obtained by the trained siamese network, $s$ is the set of images generated by DALL-E 2 with the corresponding author name $a$. In order to have $P \in [0,1]$, eq. \ref{eq:prob} is evaluated only when $min(\widehat{d}(I,s_a)) \leq 1$. Figure \ref{fig:siamese_final} shows graphically the proposed final approach

%The objective of an IP infringement metric, in our context, should be measuring how much of a protected artwork has being replicated or employed to create the new one. Indeed, Diffusion Models like DALL-E 2 do not just cut and paste from original data but they compose in a very complex manner. However, if a Resnet is able to discriminate between five authors, empirically means that is able to encode discriminating features which are typical of the authors with a certain level of composition introduced by the generative process of the DM. Given this we would like to train a metric by exploiting the resnet-18 trained previously in a siamese architecture. The training of this siamese was carried out giving as training set the same training set employed to train the resnet-18 for discrimination of author's. This time the training data was organizes in such a way to \textbf{couple similar with... couple dissimilar with... etc.} \textbf{Figure X shows the }.

%\textbf{Capire se una immagine query fa IP infringemenet devo prenderla e confrontarla con la siamese trainata con tutto il dataset da proteggere. A scopo sperimentale la confrontiamo con il dataset da proteggere (vedi definizione}

\section{Experiments and discussion}
\label{ref:results}

The siamese neural network with a baseline Resnet-18 model specialized in discriminating authors was developed for the purposes of this study as described in previous Sections. The Resnet-18 discriminative model was trained using the following parameters: ADAM optimization, 50 epochs, batch size of 32, weight decay and LR of 0.0001. The siamese configuration was trained with the loss as defined in Equation \ref{eq:loss}, ADAM optimization, 250 epochs, batch size of 64, weight decay and lr of 0.0001 and contrastive loss margin of 2.

In order to evaluate the effectiveness of the method a retrieval test was carried out. The retrieval test is able to show how the metric is working in correcting detecting the author by means of comparison between query images and all original artworks in the test set of the mixed dataset $D$.

\begin{figure}[t!]
\centering
    \includegraphics[width=\linewidth]{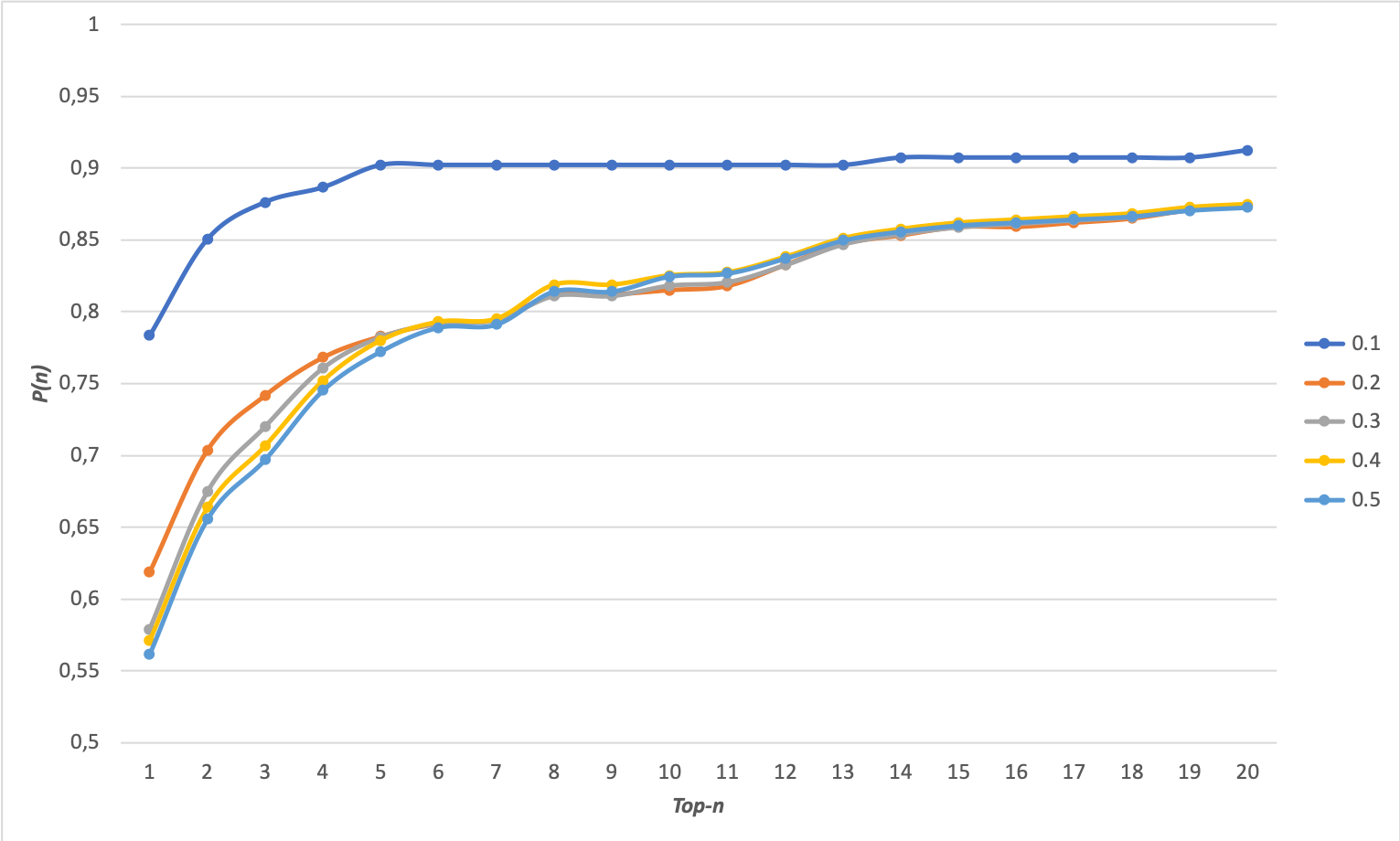}
    %\vspace{-0.5cm}
    \caption{Retrieval test results with different threshold values for the distance $\widehat{d}$ at varying of the top-n retrieved samples.}
    \label{fig:retrieval_plot}
\end{figure}

For each author a single generated image was selected as query and all the original artworks samples were employed to perform the retrieval test. 
The retrieval performance has been evaluated with the probability of the successful retrieval $P(n)$ in a number of test queries $P(n) = Q_n / Q$, where $Q_n$ is the number of successful queries according to top-$n$ criterion, i.e., the correct classification is among the first $n$ retrieved images, and $Q$ is the total number of queries. The average of $P(n)$ values with respect to all queries by considering only images achieving a distance $\widehat{d} \leq T$ where $T \in \{0.1, 0.2, 0.3, 0.4, 0.5\}$ is reported in Figure \ref{fig:retrieval_plot} at varying of $n$.

%\vspace{-0.3cm}
%\begin{equation}
%    P(n) = \frac{Q_n}{Q}
%\end{equation}

It has to be noted that images that are very close in terms of $\widehat{d}$ are not similar in terms of aspect, style, contents and semantics. This empirically implies that $\widehat{d}$ is not working on those features but learned a sort of probability of having the author name as input prompt at generation time: the only thing that unites all the artworks of an artist.
\begin{figure}[t!]
\centering
    \includegraphics[width=.8\linewidth]{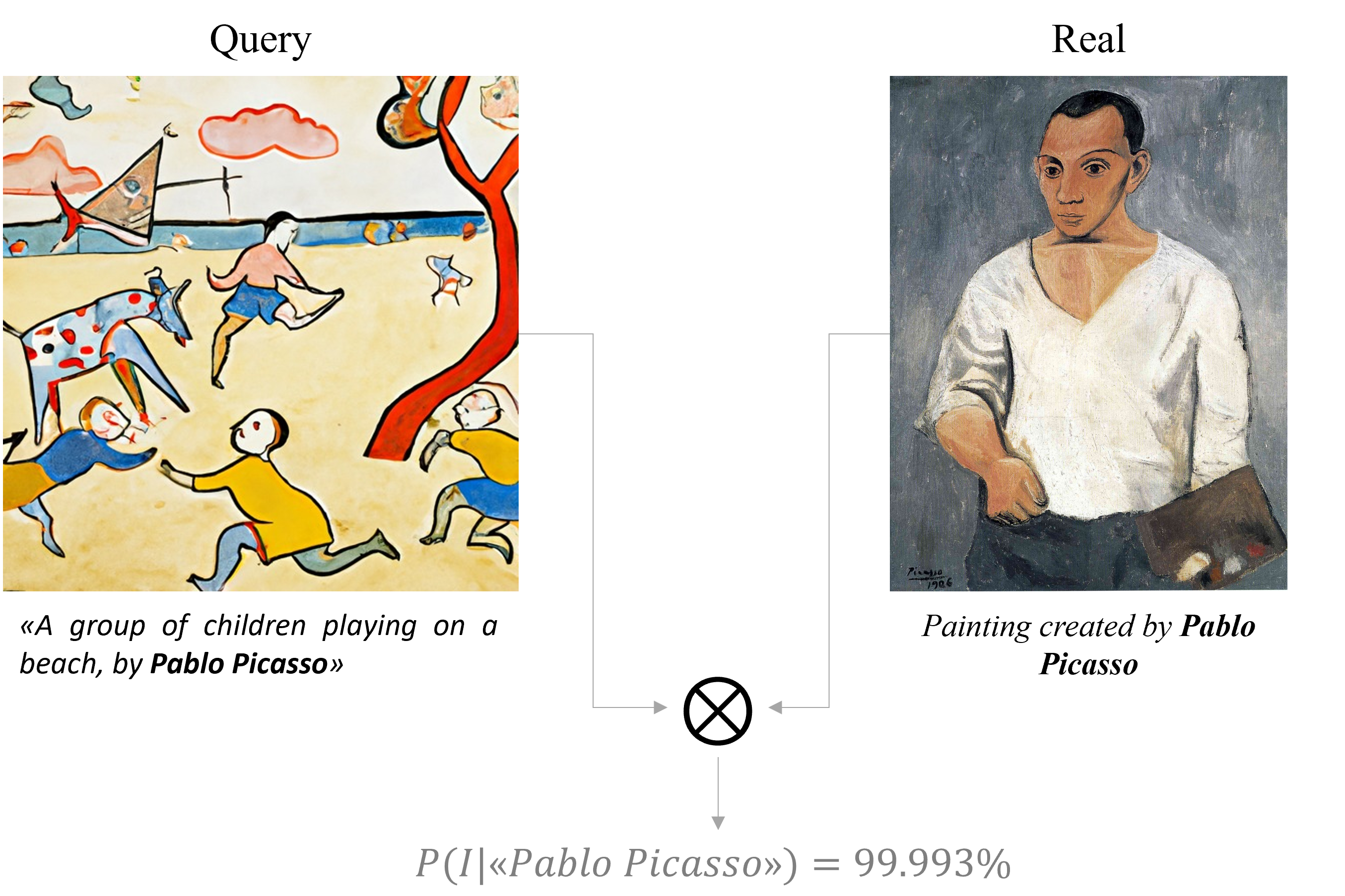}
    %\vspace{-0.5cm}
    \caption{An example with a query image $I$ (the generated one) and the real closest image found by means of the learned metric. As an overall result, a probability is obtained that the name Picasso was used for the generation of image $I$.}
    \label{fig:output_example}
\end{figure}

The results obtained demonstrate that the proposed siamese approach could be employed to protect any kind of artists database given that a re-training phase is needed to a certain amount of images generated by prompting their name. Figure~\ref{fig:output_example} shows a real case scenario.
Generalization could be achieved only by means of integrating more samples in the dataset with both original images and generated ones. Finally, it has to be noted that this work was carried out only on images generated with DALL-E limiting the generalization of the results for other DMs or tools available. While probably being robust with other images generated with DM solutions, it surely would not generalize on GANs being the already detected invisible differences extractable from images~\cite{corvi2022detection}.

%\section{Discussion on a complete solution}
%\label{ref:solution}
%TODO per il futuro paper... tocca fare un 5-fold cross validation per dimostrare la generalizzazione del sistema e 
%In sintesi possiamo riepilogare come mettere su un sistema.

%1) Un autore viene da te e ti chiede di proteggere i suoi contenuti multimediali (ad es. artworks).
%2) Tu metti i suoi contenuti multimediali nel reference dataset to be protected 
%3) Un sistema di monitoraggio dovrebbe osservare le immagini generate e che girano ad esempio sui social o vengono vendute e vengono prese un set di immagini di query
%4) Ogni immagine di query viene confrontata con la siamese metric con tutto il dataset e se la % di similarità è superiore al 50% allora un allarme viene generato

\section{Conclusion and future works}
\label{ref:conclusion}
In this paper, a novel approach to estimate the probability of an artist's name being used in the input prompt of an image generated by a diffusion model, specifically DALL-E 2, was presented. Our approach aimed to address the concern of potential intellectual property infringement in the context of image generation, as the usage of an artist's name in the input string might imply that the diffusion model has learned from some or all of the artist's work, potentially violating their copyright.

This work employed metric learning for classification of an extremely limited number of authors but in future more sophisticated similarity measures, larger and more diverse datasets, and additional techniques to refine the estimation of the input string's content could be explored. Additionally, investigating the ethical implications and possible solutions to the challenges posed by AI-generated content in relation to copyright and intellectual property protection should be a priority for the research community.

 \bibliographystyle{splncs04}
 \bibliography{references}
\end{document}